\pdfoutput=1

\documentclass[11pt]{article}

\usepackage[]{ACL2023}

\usepackage{times}
\usepackage{latexsym}

\usepackage[T1]{fontenc}

\usepackage[utf8]{inputenc}

\usepackage{microtype}

\usepackage{inconsolata}

\usepackage[utf8]{inputenc}

\usepackage{microtype}
\usepackage{graphicx}
\usepackage{amsmath}
\usepackage{amssymb}
\usepackage{booktabs}
\usepackage{comment}
\usepackage{xspace}

\usepackage{xcolor}

\newcommand{\model}{\textsc{LMCap}\xspace}
\newcommand{\sms}{Socratic Models\xspace}

\title{\model: Few-shot Multilingual Image Captioning by \\Retrieval Augmented Language Model Prompting}

\author{Rita Ramos$^{\dagger}$ \ \ Bruno Martins$^{\dagger}$ \ \ Desmond Elliott$^{\star}$ \\
        $^{\dagger}$INESC-ID, Instituto Superior Técnico, University of Lisbon \\ $^{\star}$Department of Computer Science, University of Copenhagen \\
        \texttt{ritaparadaramos@tecnico.ulisboa.pt}}

\begin{document}

\maketitle

\begin{abstract}

Multilingual image captioning has recently been tackled by training with large-scale machine translated data, which is an expensive, noisy, and time-consuming process. Without requiring any multilingual caption data, we propose \model, an \textit{image-blind} few-shot multilingual captioning model that works by prompting a language model with retrieved captions. Specifically, instead of following the standard encoder-decoder paradigm, given an image, \model first retrieves the captions of similar images using a multilingual CLIP encoder. These captions are then combined into a prompt for an XGLM decoder, in order to generate captions in the desired language. In other words, the generation model does not directly process the image, instead processing retrieved captions. Experiments on the XM3600 dataset of geographically diverse images show that our model is competitive with fully-supervised multilingual captioning models, without requiring any supervised training on any captioning data.
\end{abstract}

\section{Introduction}

The task of image captioning has witnessed impressive performance gains with the trend of large-scale encoder-decoder models and vision-and-language pre-training \cite{li2022blip,wang2021simvlm, hu2022scaling, wang2022git}. Despite all of this progress, existing models are mostly available on English or are specialised for other high-resource languages. This limits the access to the technology for a broader range of languages that exist in the world. Moreover, the current mainstream trend results in design decisions and methods that may only work well for English-centric datasets or the couple of languages for which captioning data is available \cite{ruder2020beyondenglish}. There is a need to develop multilingual image captioning models that can serve speakers of different languages.

Still, scaling captioning models to a wide variety of languages involves different challenges. One major limitation is the lack of multilingual image-caption pairs of clean labelled data for training the models. One possible solution is to automatically translate the existing English datasets \cite{thapliyal2022crossmodal}. While effective, this approach can result in models that learn translation artefacts, and perpetuates an English-centric perspective instead of encouraging the use of geographically diverse concepts that are not overly specific to the western culture \cite{liu2021visually}. Moreover, with or without automatic translations, training captioning models with multilingual data can be expensive, given the amount of data and number of parameters needed to mitigate the \emph{curse of multilinguality}~\cite{conneau2019unsupervised, goyal2021larger}. 

This paper presents \model, an \textit{image-blind} multilingual image captioning model that does not require any training specific for image captioning. We propose an efficient method that reuses a pre-trained multilingual language model and adapts it to the vision-and-language captioning setting. Our work is motivated by the recent "\sms{}" framework \cite{zeng2022socratic}, in which different models can be combined through text prompting (e.g., image captioning can be achieved by prompting a language model with a set of visual concepts extracted from the predictions of a vision model). Different from the original Socratic Models, our approach is inspired by retrieval-augmented generation \cite{rag,izacard2022few}. Specifically, a multilingual language model generates captions given a prompt consisting of the captions retrieved from similar images, and a demonstration of how to produce a caption in the desired language. We note here that this is an \textit{image-blind} approach, i.e. the language model producing the caption does not actually process the image.

Our main contributions are as follows: (1) We propose a few-shot multilingual image captioning approach named \model, that re-uses pre-trained models without requiring any training specific for image captioning; (2) To the best of our knowledge, \model is the first captioning model with retrieval-augmented generation in a multilingual setting, and in a few-shot setting of captioning;
(3) We report on experiments with the XM3600 benchmark \cite{thapliyal2022crossmodal} of human-authored captions and geographic diverse images, demonstrating that \model exhibits strong few-shot performance on a
wide variety of languages; %
(4) We further show that \model performs substantially better than the original Socratic Models. Moreover, instead of only achieving competitive performance against other zero-shot models, \model can also compete with a large-scale supervised state-of-art captioning model.

\section{Background and Related Work}

\paragraph{Image Captioning:} The task of automatically generating textual descriptions for input images has been largely explored in English, while multilingual image captioning has only been addressed in a couple of studies \cite{gu2018unpaired,thapliyal2022crossmodal,chen2022pali}. Like in most recent work on image captioning \cite{li2022blip,wang2021simvlm, wang2022git}, studies addressing multilingual setups have also focused on scaling the size of encoder-decoder models and the amount of training data, resorting to machine translated versions of multimodal data to accommodate multiple languages \cite{thapliyal2022crossmodal}. Differently from training a large-scale encoder-decoder model, we follow a few-shot setting with an \textit{image-blind} approach based on prompting.

\paragraph{Few-Shot and Zero-Shot Approaches:} Performing few-shot learning by prompting a language model with examples and demonstrations of a task \cite{brown2020language, Radford2019LanguageMA, schick2020exploiting} is an efficient and effective alternative to update model parameters. Similarly to other NLP tasks, recent work in the vision-and-language domain has used prompt-based learning by building on top of pre-trained language and vision models, although usually also involving extra multimodal training \cite{frozen,alayrac2022flamingo, jin2021good}. In our work, we follow a similar few-shot prompting approach to the recent \sms{} \cite{zeng2022socratic} that do not involve any multimodal training, as described next. In image captioning, there have also been zero-shot methods that similarly to our approach do not involve any training, by relying on prompts or adaptations over the decoding algorithms, such as ZeroCap \cite{tewel2021zero} and ConZic \cite{zeng2023conzic}. However, these models work for English and not for the multilingual captioning setting. 

\paragraph{Socratic Models:} \citet{zeng2022socratic} proposed the Socratic Models (SMs) framework, where different multimodal pre-trained models communicate via zero-shot or few-shot prompting. For the task of image captioning, SMs generate captions by prompting a language model (i.e., GPT-3 \cite{brown2020language}) with information about the input image obtained with another pre-trained model (i.e., CLIP \cite{radford2021learning}). The visual information is in this way represented into a language-based prompt, containing the number of people presented in the image, the places, objects, and what is the type of image. We explore a similar approach in the multilingual setting by reusing multilingual models, and through a retrieval-based prompt.

\paragraph{Retrieval-augmentation:} The knowledge from language models can be adapted and expanded by combing non-parametric knowledge from datastores (i.e., external memories) \cite{khandelwal2019generalization, rag, izacard2022few, ram2023context}. The success of conditioning generation with retrieved information, in several different NLP tasks, has inspired some recent studies in image captioning \cite{ramos-etal-2023-retrieval, fei2021memory, sarto2022retrieval, ramos2023smallcap}. The study that is most closely related to our captioning model is SmallCap \cite{ramos2023smallcap}, an encoder-decoder model that is prompted with retrieved captions as well. However, in image captioning, retrieval-augmentation has mostly being explored with supervised learning and not few-shot learning. Moreover, retrieval-augmentation remains unexplored in the multilingual scenario.

\section{Model}

\begin{figure*}[!ht]
    \centering
    \includegraphics[width=\linewidth]{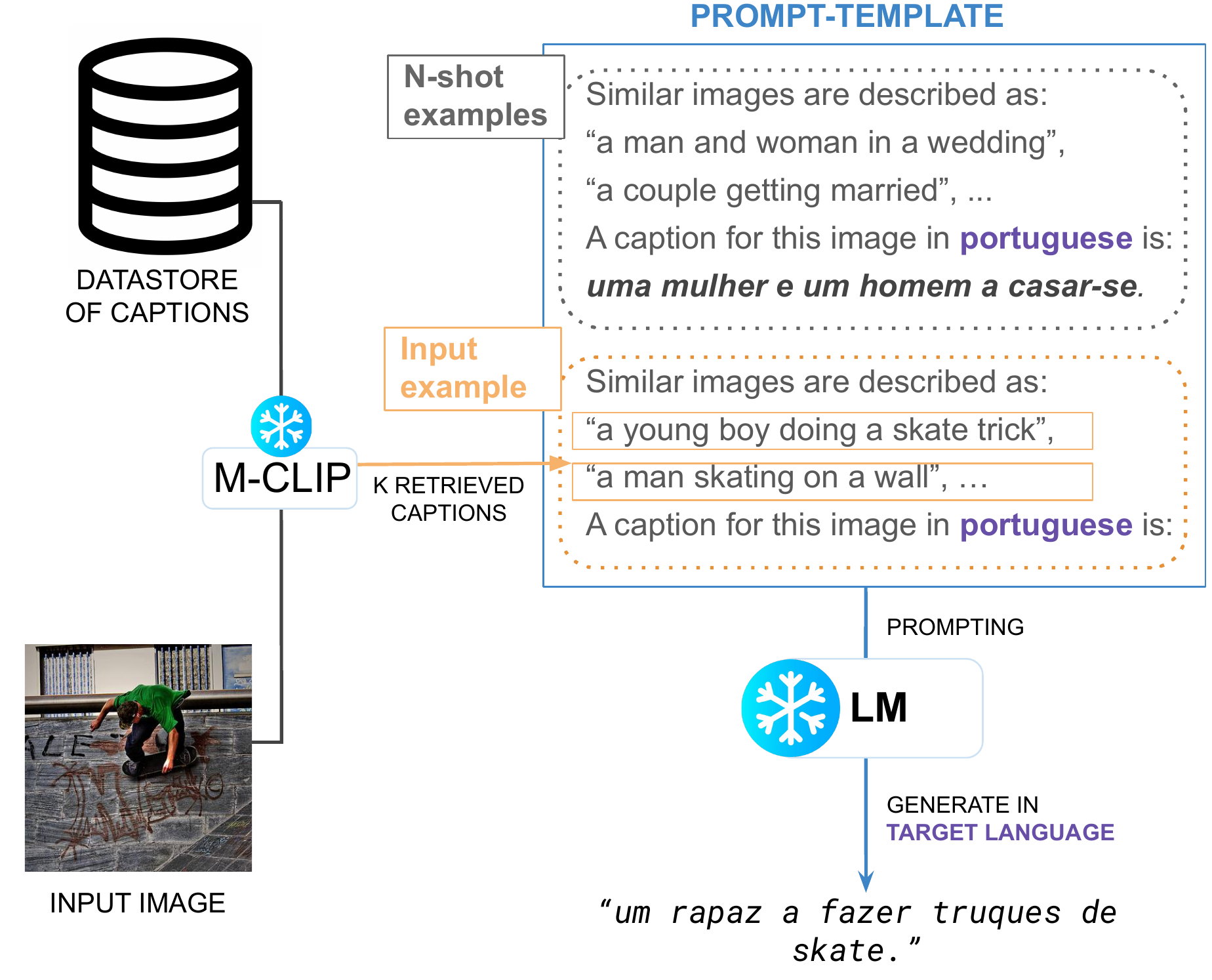}
    \caption{Illustration of the key aspects of \model, a few-shot multilingual image captioning
approach that re-uses pre-trained unimodal
models without requiring any training. In our \textit{image-blind} approach, a multilingual language model (XGLM) is prompted with information retrieved with a multilingual CLIP model. The prompt contains a set of $N$-shot examples and $K$ retrieved captions, to guide caption generation in a desired language.}
    \label{fig:approach}
\end{figure*}

Language Model Prompt-based Captioning (\model) is a few-shot multilingual captioning model augmented with retrieval. It involves prompting a Language Model (LM) with captions retrieved from a datastore by a Vision-and-Language Model (VLM). Captions are generated in an \textit{image-blind} manner, without actually processing the visual contents of the input image, instead using a prompt containing the retrieved captions. The method works as follows: first, given an input image, the VLM is used to find relevant captions in the datastore. Second, the retrieved captions are converted to a language prompt, which is encoded by the multilingual LM to generate captions in a desired language, conditioning the generation on the prompt. Finally, the set of generated captions can be scored by the VLM against the input image, to select the best caption. The main aspects of our approach are shown in Figure \ref{fig:approach} and fully detailed next.

\paragraph{Image-Text Retrieval:} The input image and a datastore of captions are encoded by a multilingual CLIP \cite{carlsson}, i.e. a VLM that can be used to calculate image-text similarity. In this way, given the encoded data, M-CLIP is used to retrieve the $K$ most similar captions from the datastore. The datastore contains captions associated to diverse images, which can be in English or another language. The retrieved captions will serve to guide a language model as an example of what the predicted caption should resemble, through the use of a prompt and as described next.

\paragraph{Retrieval-augmented Prompting:} The retrieved captions, which represent the visual information about the image, are formatted into a prompt for the language model. The prompt starts with fixed $N$-shot examples and ends with the retrieval information about the input image, to guide the language model. Each shot is a demonstration of how to generate a caption in a desired language for an image, given a set of retrieved captions. After these $N$-examples, the prompt terminates with the retrieved information about the actual input image. An example of %
the format of the prompt can be seen in Figure \ref{fig:approach} and in more detail in Appendix \ref{app:prompt}. We note that the retrieved captions, either from the fixed $N$-shot examples or those corresponding to the input image, can be presented in any language or in multiple languages.

\paragraph{Prompting Multilingual Text Generation:} The aforementioned prompt is used as input for an XGLM \cite{lin2021few} pre-trained multilingual autoregressive LM, to generate captions in a given language. XGLM is applied in a few-shot setting, which means that \model does not require any training (i.e., the captions are generated by providing the prompt at inference time to XGLM). Captions are generated in the desired language by including an example in the $N$ demonstrations in the prompt, as shown in Figure \ref{fig:approach}.

\paragraph{Multilingual Reranking:} After the LM generates a set of captions, the multilingual VLM performs a final image--text similarity step to find the caption that best describes the input image. This is based on the same M-CLIP model used for the initial image--text retrieval.

\section{Evaluation}

In this section, we describe the evaluation of \model. We describe the experimental setup and results, and we also present ablation studies and further discussions about our approach. 

\subsection{Experimental Setup}

\paragraph{Model:} \model uses two pre-trained multilingual models, namely the autoregressive XGLM language model  \texttt{facebook/xglm-2.9B}, and the multilingual M-CLIP vison-and-language model \texttt{xlm-roberta-large-ViT-H-14}, respectively available on HuggingFace~\cite{wolf2020transformers} and OpenCLIP\footnote{\url{https://github.com/mlfoundations/open_clip}}. Our approach does not require any training, generating captions at inference time using a single NVIDIA V100S 32GB GPU. 

To generate a caption in a desired language, XGLM is prompted with retrieved captions extracted by the M-CLIP model. For caption retrieval, the input image and a set of captions from a datastore are both encoded by M-CLIP to perform direct image-text search. The datastore contains English captions from the COCO training set and is indexed offline with the nearest-neighbour search library named FAISS \cite{JDH17}, using the index \texttt{IndexFlatIP} that does not involve training. A set of $K$=$4$ retrieved captions are used in the prompt for the input image, along with a fixed set of $N$=$3$-shot examples, as described in Appendix \ref{app:prompt}. Conditioned on the prompt, XGLM generates captions using beam-search decoding with a beam of 3. A set of $c$=$3$ candidate captions are re-ranked using M-CLIP, to select the final generated caption in the desired language. The code for \model is made freely available\footnote{\url{https://github.com/RitaRamo/lmcap}}.

\paragraph{Datasets:} We mainly evaluate our approach on XM3600, i.e. a multilingual image captioning dataset \cite{thapliyal2022crossmodal} featuring geographically-diverse images, collected from Open Images with basis on the regions of 36 languages. For each language, 100 images were selected and annotated with human generated captions, resulting in a total of 3600 images and 261375 captions across the 36 languages. XM3600 does not contain training or validation splits.

For validation and hyperparameter tuning, we relied on the COCO \cite{chen2015microsoft} validation split (COCO-DEV) from the standard Karpathy splits \cite{karpathy2015deep}. For ``reference captions'', we machine translate the English captions into Spanish, Hindi, and Chinese, using the M2M-100 model \cite{fan2021beyond}, similarly in spirit to \citet{thapliyal2022crossmodal} who used the Google Translate API\footnote{\url{https://cloud.google.com/translate}}. We make this development set available to the community at \url{https://github.com/RitaRamo/lmcap}. As previously mentioned, we also use the captions from the COCO training set to build the datastore. The datastore simply contains the original English captions from COCO without incurring in an expensive and noisy machine translation process, unlike in the study from \citet{thapliyal2022crossmodal}. 

\paragraph{Model Assessment and Comparison:}

We compare \model with the four multilingual models proposed by \citet{thapliyal2022crossmodal}. These models combine different mT5 \cite{xue2020mt5} and ViT \cite{zhai2022scaling} versions and are trained in a fully-supervised fashion on COCO-35L and CC3M-35L, i.e., Google’s machine translation API versions of the original COCO and CC3M datasets \cite{chen2015microsoft, sharma2018conceptual}. Specifically, \texttt{BB+CC} combines mT5-base and ViT-B/16 pretrained on CC3M-35L and finetuned on COCO-35L; \texttt{BB} is trained on COCO-35L; \texttt{Bg} switches to the ViT-g/14 model; and \texttt{Lg} uses mT5-large and and ViT-g/14, also trained with COCO-35L. For reference, \citet{thapliyal2022crossmodal} spent 5000 TPU hours to train their models, while our method can be used out-of-the-box for inference, i.e., 45 minutes for the X3600 benchmark per language.

Following \citet{thapliyal2022crossmodal}, results are reported with the CIDEr \cite{vedantam2015cider} metric for English, Spanish, Hindi, and Chinese, with other languages covered in Section \ref{disc:36lang}. CIDEr is a standard captioning metric that computes how well the generated caption matches the consensus of the reference captions, based on Term Frequency–Inverse Document Frequency (TF-IDF). In Appendix \ref{app:eval_metrics}, we included more generation metrics for holistic evaluation. To compute the metrics, we used the COCO evaluation package \footnote{Available at \url{https://github.com/tylin/coco-caption}}, and the SacreBLEU tokenization \cite{post-2018-call}. 

\subsection{Results}

\paragraph{XM3600:} Following \citet{thapliyal2022crossmodal}, we report results on XM3600 for English, Spanish, Hindi, and Chinese, in Table \ref{tab:xm3600}. %
We can see that \model outperforms all supervised approaches on Chinese, and achieves competitive performance on the other languages, despite being \textit{image-blind} and not being trained on any image captioning data. For English, Spanish, and Hindi, we note that \model is only outperformed by the large-scale supervised variant BB+CC, pre-trained on CCM3 and fine-tuned on COCO, jointly on English and the other 35 languages for the two datasets, i.e., with 123M captions. For the other variants that are only trained on COCO-35L, our model has a substantially larger performance on the CIDER metric across all four languages. We also show that our model can further benefit from increasing the datastore (\model{}$_{+}$), %
as described in more detail over Section \ref{ablation}. %

 \begin{table}[h!]
    \resizebox{\linewidth}{!}{
    \centering
    \begin{tabular}{lllll}
    \toprule
        Model & en & es & hi & zh   \\ 
 \midrule    
         \multicolumn{5}{c}{Multilingual Captioning Supervised Learning} \\ 
      
         \midrule  
        \emph{BB+CC} &  \textbf{0.584} & \textbf{0.425} & \textbf{0.197} & 0.202  \\ 
      
        \emph{BB} & 0.297 &0.194 & 0.098 & 0.087  \\
        \emph{Bg} &  0.337 & 0.232 & 0.112 & 0.110\\ 
        \emph{Lg} &  0.343  &0.220& 0.111&  0.099\\ 
        \midrule   
         \multicolumn{5}{c}{Few-shot Learning} \\ 
         \midrule    
        \model  & 0.452  &  \underline{0.329}  &  \underline{0.132} &   \underline{0.221} \\
        \model{}$_{+}$ & \underline{0.526}  &  0.326  &  0.078 &   \textbf{0.251} \\
    \bottomrule
    \end{tabular}
    }
    \caption{Results on the geographically-diverse XM3600 benchmark. We compare our 
few-shot \model model against large-scale supervised multilingual and multimodal SOTA models proposed by \citet{thapliyal2022crossmodal}. Best results in bold and second-best underlined. %
    }
    \label{tab:xm3600}
\end{table}

\paragraph{COCO:} For completeness, we also report results on the machine translated COCO-DEV set in Table \ref{tab:coco}. In the top half of the table we show the performance of the 4 SOTA models on COCO-DEV via Google’s machine translation API. Since this dataset was not provided by the authors, we perform as well automatic machine-translation but using the M2M-100 model \cite{fan2021beyond}, which gives an approximation for model comparison on COCO. As expected, \model is outperformed on COCO since all the 4 variants were trained on it across 36 languages, with a large number of trainable parameters. Our model still reaches impressive performance, considering it was not trained on COCO for any of those languages, neither was it trained on any multimodal data. This is especially the case for English, where our model reaches a similar CIDEr score, although it only reaches about half the performance for the other languages. In Appendix \ref{app:more_coco}, we also compare \model with prompt-based captioning methods that were specially designed for English.

\begin{table}[!h]
    \resizebox{\linewidth}{!}{
    \centering
    \begin{tabular}{lcllll}
    \toprule
        Model & $|\theta|$ & en & es & hi & zh   \\ 
        \midrule    
        \multicolumn{6}{c}{COCO-DEV-GOOGLE} \\ 
        \midrule    

        BB+CC & 766 &  0.980 & 0.962 &0.759 & 0.748  \\ %
        BB & 1230 & 0.856 & 0.844 & 0.671 & 0.659  \\ %
        Bg & 1691 & 0.851 & 0.835 & 0.718 & 0.695\\ 
        Lg & 2241 &  0.875 &0.859& 0.624 &  0.656\\ 
         \midrule    
        \multicolumn{6}{c}{COCO-DEV-M2M100} \\ 
         \midrule    
        \model & N/A & 0.767 &  0.453 & 0.334 & 0.584  \\

    \bottomrule
    \end{tabular}
    }
    \caption{CIDEr performance on the COCO dataset. The top of the table presents SOTA results on the COCO validation split, translated via the GOOGLE API. The bottom rows of the table shows our model performance on COCO-DEV, translated via the M2M-100 model. $|\theta|$ corresponds to the number of trainable parameters in the model (in millions). %
    }
    \label{tab:coco}
\end{table}

\subsection{Ablation Studies}
\label{ablation}

To better understand the design choices of \model, we report a series of ablation tests on COCO-DEV, to avoid direct tuning on the XM3600 benchmark. 

\paragraph{Prompt:} \label{para:prompt} Given that \model works by prompting a language model with $K$ retrieved captions and $N$-shot examples, we study the effect of our prompt when varying $K$ and $N$. Table \ref{tab:k_n_vary} shows the importance of not depending on a single retrieved caption across the 4 languages. This is similar to previous findings in retrieval-augmentated captioning studies focusing on English \cite{sarto2022retrieval,ramos2023smallcap}, which showed that a large $K$ makes the model more robust to mismatched captions. We further see that English and Spanish benefit from encoding a larger set of retrieved captions, while Hindi and Chinese work better with a smaller $K$. We select $K=4$ since it has close-to-optimal performance for each of the languages. We then explore varying the number of $N$-shot examples, and found $N=3$ to be the optimal value on all the four the languages. We thus use $K=4$ and $N=3$ in the prompt of \model.

\begin{table}[!ht]
    \resizebox{\linewidth}{!}{
    \centering
    \begin{tabular}{lllll}
    \toprule
        Setup & en & es & hi & zh \\ 
         \midrule    
        \multicolumn{5}{c}{Varying K-Captions} \\ 
         \midrule
        K=1, N=1 &  0.622 & 0.380 & 0.240 & 0.522  \\ %
        K=2, N=1 &  0.654 &0.400 &\textbf{0.269} &  0.562\\ %
        K=3, N=1 &  0.695 & 0.414 & 0.211 & \textbf{0.565} \\ %
        K=4, N=1 &  0.711 & 0.415 & 0.229 & 0.554  \\ %
         K=5, N=1 & \textbf{0.734} & \textbf{0.424} & 0.205 & 0.529  \\ %
        \midrule    
        \multicolumn{5}{c}{Varying N-Shot} \\ 
        \midrule    

          K=4, N=1 &  0.711 & 0.415 & 0.229 & 0.554 \\ %
         K=4, N=2 & 0.735 & 0.440 & 0.247 & 0.583 \\ %
        K=4, N=3 &  \textbf{0.767} & \textbf{0.454} & \textbf{0.334} & \textbf{0.584} \\ %
        K=4, N=4 & 0.757 & 0.424 & 0.318 & 0.580 \\ %
    \bottomrule
    \end{tabular}
    }
    \caption{The effect of using different numbers of $K$ retrieved captions and $N$ few-shot examples. Results reported on COCO-DEV with best results in bold.
    }
    \label{tab:k_n_vary}
\end{table}

\paragraph{Datastore:} \label{para:datastore} We also studied different contents for the datastore beyond the English captions from the COCO training set, shown in Table \ref{tab:datastores}. Given that our model reaches much better performance on English, we hypothesise that our model can better generate captions in a desired language when having the retrieved captions in that same language. This could be validated using translations from COCO in the other languages, but since those are not available, we instead used a machine translated version of the Conceptual Captions dataset (CCM3) from \citet{qiu2022multilingual}. We used the English, Spanish, and Chinese versions of the CCM3 training set, respectively for each of the corresponding languages (CCM3-L). We found that performance deteriorates on the COCO-DEV dataset, which might be explained by the difference between the COCO and CCM3-L datasets. Even combining the two datasets (COCO + CCM3-L) is worse than using only the COCO dataset. 

In an attempt to cover more diverse concepts, we augmented COCO with three large web datasets (Conceptual Captions \cite{sharma2018conceptual}, Conceptual 12M \cite{changpinyo2021conceptual}, and SBU captions \cite{ordonez2011im2text}), using their noise-free versions~\cite{li2022blip}. We refer to this dataset as CCS, and it contains synthetic model-generated texts for the web images. Using CCS leads to an improvement compared to just using COCO, except for Hindi. In Table \ref{tab:xm3600}, we also report results on XM3600 with this best datastore configuration, for which the performance again decreases for Hindi, but has a substantial improvement on English and Chinese. The benefits of including a more diverse collection of captions are further shown in Apprendix \ref{app:datastore_Examples} with some qualitative examples (e.g., \model was now able to generate the french concept \textit{macarons} in English). Notice that the retrieved captions from CCS are still on English. Thus, although there is lack of multilingual image-caption pairs with clean labelled data, it would be interesting to pursue further work on incorporating retrieved information from other languages, in order to improve performance to levels similar to those for English.

\begin{table}[!ht]
    \resizebox{\linewidth}{!}{
    \centering
    \begin{tabular}{lllll}
    \toprule
        Datastores & en & es & hi & zh   \\ 
        \midrule    
        COCO &  0.711 & 0.415 & 0.229 & 0.554  \\ %
        CC3M-L &  0.387 & 0.309 & - & 0.337  \\ %
        COCO + CC3M-L &  0.601 & 0.359 & -& 0.481  \\ %
        COCO + CCS & \textbf{0.713} & \textbf{0.431} & 0.212 & \textbf{0.563}  \\ %
    \bottomrule
    \end{tabular}
    }
    \caption{Datastore ablations on COCO-DEV, where captions are retrieved from different sources of data. CC3M-L corresponds to machine translated version of Conceptual Captions proposed in \citet{qiu2022multilingual} (Hindi not available), while CCS refers to the Conceptual Captions, Conceptual 12M, and SBU datasets \cite{li2022blip}.
    }
    \label{tab:datastores}
\end{table}

\paragraph{Model Size:}\label{para:model_size}

In Table \ref{tab:model_size}, we show the importance of using a language model that has a sufficiently large number of parameters. Both XGLM-562M and XGLM-1.7B are unable to generate captions beyond English. On the other hand, the 7.5B variant can lead to a stronger performance, but large-scale LMs require more GPU memory, which limits the size of the prompt that can be encoded with modest hardware\footnote{We had to run the largest model in half precision (float16).}. \model uses the more efficient XGLM-2.9B version. These results are in
line with previous findings, which suggest that stronger few-shot performance is achieved when the prompt is encoded by large LMs \cite{brown2020language}.

\begin{table}[!ht]
    \resizebox{\linewidth}{!}{
    \centering
    \begin{tabular}{lllllll}
    \toprule
        Params & Config. & RAM & en & es & hi & zh   \\ 
        \midrule    

        564M & K=4, N=3 & 6G & 0.411 & 0.094  & 0.030 &  0.146\\ %
        1.7B & K=4, N=3 & 12G & 0.637 & 0.143 &  0.066 &  0.272  \\ %
        2.9B & K=4, N=3 & 16G & 0.767 & 0.454 & 0.334 & 0.584  \\ %
        7.5B & K=4, N=3 & 22G & \textbf{0.787}  & \textbf{0.489} & \textbf{0.365} & \textbf{0.644}  \\ %
        
    \bottomrule
    \end{tabular}
    }
    \caption{CIDEr performance on COCO-DEV, across the different variants of XGLM, to show the scaling behaviour of the LM used in \model. RAM corresponds to the GPU memory
consumption.
    }
    \label{tab:model_size}
\end{table}

\subsection{Additional Discussion}
We now discuss the performance of \model across the 36 languages, taking into consideration the data that was used for pre-training the LM. We also compare our approach with SMs and a simple baseline of retrieval plus translation. To support quantitative evaluation, we show some qualitative examples.

\paragraph{Multilingual Pre-training:} \label{disc:36lang} In Table \ref{tab:all_languages}, we report the results of \model on XM3600 for all the 36 languages considered in the dataset, ordered by the percentage of pre-training data used in XGLM for each language. \model shows strong few shot performance on the diverse set of languages in which XGLM was pre-trained on. Similarly to BB+CC and Lg models, which are limited to the 36 languages they were trained on, our model is also dependent on the LM pre-training data, although there is potential to replace XGLM by another large LM, in order to generalize to other languages.

\begin{figure*}[!ht]
    \centering
    \includegraphics[width=\linewidth]{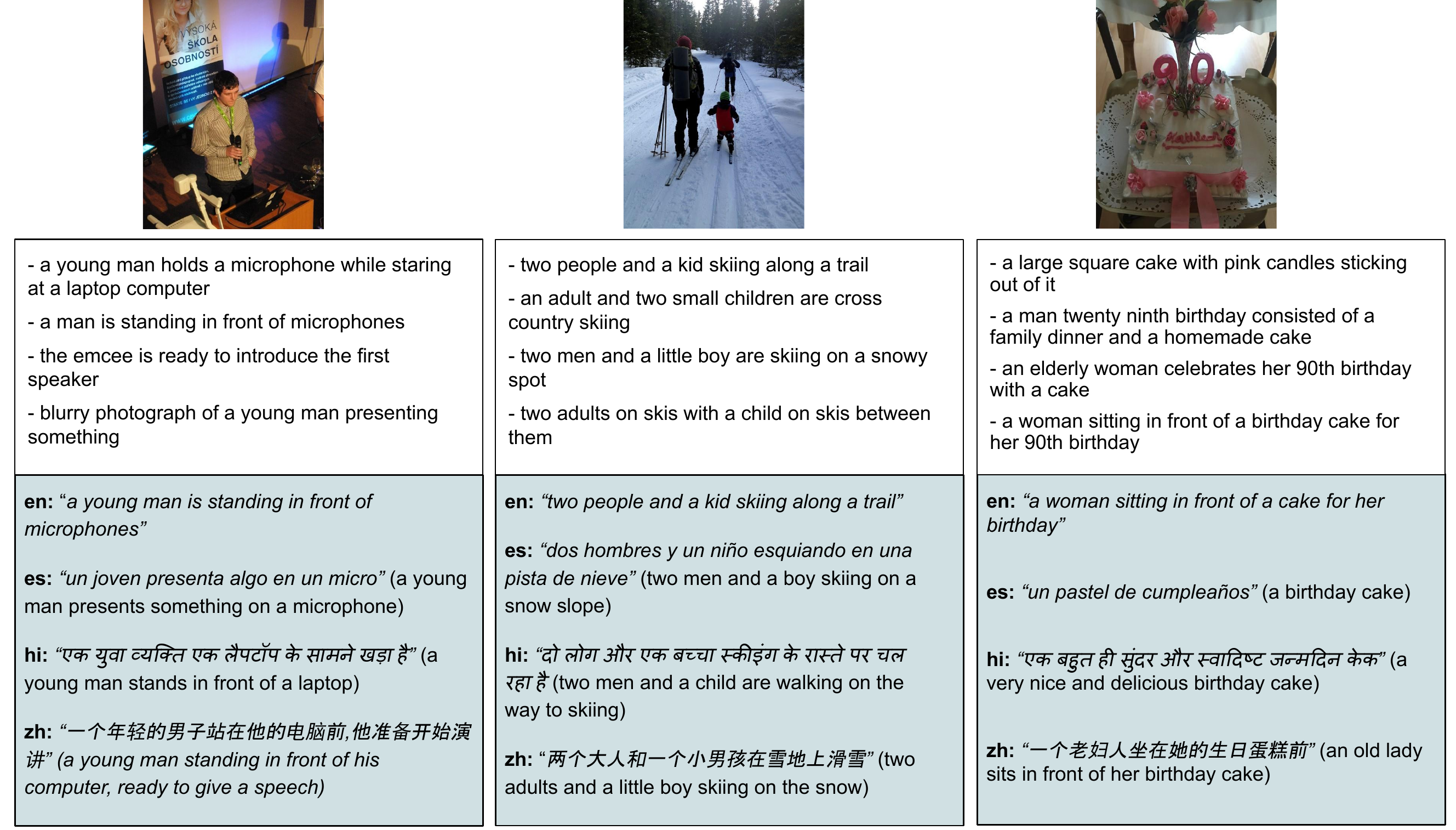}
    \caption{Examples of captions generated by \model for English, Spanish, Hindi, and Chinese,   
    on XM3600 images and based on retrieved captions regarding each \textit{blind-input} image.
    }
    \label{fig:examples}
\end{figure*}

\begin{table}[t!]
    \centering
    \begin{tabular}{lccc}
    \toprule
        & \textsc{BB+CC} &  \textsc{Lg}  & \model   \\ 
        \midrule    
        en &  0.584  & 0.343 & 0.452 \\
        ru & 0.194  & 0.089 &  0.134 \\
        zh & 0.202 & 0.099 &  0.221\\
        de &  0.224   & 0.130 &   0.153\\ %
        es &  0.425 & 0.220 &  0.329\\
        fr&  0.410 & 0.217&  0.260\\
        ja & 0.254  & 0.141 & 0.161 \\
        it & 0.321  & 0.168 &  0.226 \\
        pt & 0.380   & 0.202 & 0.283 \\
        el &  0.199   & 0.101 &   0.136\\
        ko & 0.288  & 0.152 & 0.157 \\
        fi &  0.177  & 0.089 &  0.112 \\
        id & 0.307  & 0.167 & 0.151 \\
        tr & 0.232   & 0.122 & 0.103 \\
        ar &  0.227 &  0.106 & 0.107   \\ %
        vi & 0.336   & 0.182 & 0.265 \\
        th & 0.418  & 0.226 &  0.166 \\
        hi & 0.197  &0.111 &  0.132\\
        bn &  0.200 &  0.133 & 0.022  \\ %
        sw & 0.319  & 0.151 &  0.085 \\
        te & 0.196  & 0.099 &  0.042 \\
        \midrule
        \multicolumn{4}{l}{\textit{Languages not in XGLM pretraining data}} \\
        \midrule    
        cs  &  0.313 & 0.139 & 0.005  \\ %
        da &  0.329   & 0.192 & 0.020 \\ %
        fa &  0.311  & 0.155 &  0.002 \\
        he &0.230   &0.098 & 0.001 \\
        hr & 0.224 & 0.085 &  0.001\\
        hu & 0.175&  0.096 & 0.006 \\
        mi & 0.405 & 0.243 & 0.015 \\
        nl & 0.441    & 0.232 &  0.082  \\
        no & 0.385   & 0.230 & 0.025 \\
        pl & 0.236   & 0.108 & 0.003 \\
        ro & 0.188  & 0.100 &  0.007 \\
        sv & 0.370  & 0.225 &  0.077 \\
        uk & 0.189   & 0.081 &   0.006 \\
        \midrule    
        AVG$^{*}$ & 0.290 & 0.154 & 0.176 \\
    \bottomrule
    \end{tabular}
    
    \caption{Results for the 36 languages on the XM3600 benchmark, ordered by the pre-training language ratio of XGLM. BB+CC and LG are full-supervised state-of-the-art approaches from \citet{thapliyal2022crossmodal}. AVG$^{*}$ corresponds to the average performance across the languages on which XGLM was pre-trained. 
    }
    \label{tab:all_languages}
\end{table}

\paragraph{Comparision with Socratic Models:} Since \model is inspired in Socratic Models (SMs), we compare them against our approach. For this, XGLM receives the Socratic prompt that includes the image type, the number of people, places and object categories\footnote{Using the original code at \url{https://colab.research.google.com/drive/1KOlc9nN0NJ5GAif_dmuOqsRqlZycoIrc?usp=sharing}}, instead of our retrieved captions. Results are reported in Table \ref{tab:socratic}. Compared to either zero-shot or few-shot SMs, we can see that our model largely outperforms SMs, with a noteworthy CIDER improvement of more than 39.1\% on English, 20.0\% on Spanish, 11.5\% on Hindi, and of 21.4\% Chinese. This confirms the effectiveness of our retrieval-augmented LM prompting approach.

\begin{table}[!h]
    \resizebox{\linewidth}{!}{
    \centering
    \begin{tabular}{lllll}
    \toprule
        Model & en & es & hi & zh   \\ 
        \midrule    

        Socratic &  0.067 & 0.045 & 0.001 & 0.031 \\ %
        Socratic N=1 &  0.454 & 0.280 & 0.176 & 0.340  \\ %
        Socratic N=2 &  0.344 & 0.215 & 0.141 & 0.268 \\ %
        Socratic N=3  &0.376 & 0.254 & 0.219 & 0.370 \\ %
         \model &  \textbf{0.767} & \textbf{0.454} & \textbf{0.334} & \textbf{0.584} \\ %

    \bottomrule
    \end{tabular}
    }
    \caption{Comparison to Socratic Models (SMs) on the COCO-DEV dataset. \model clearly outperforms SMs, as highlighted by bold.
    }
    \label{tab:socratic}
\end{table}

\paragraph{Baseline of Retrieval with Translation:} We also compared our approach against a baseline that retrieves the nearest caption on English and translates it into other languages in Table \ref{tab:baseline}, using the M2M-100 model. This is to quantify the impact of prompting the language model compared to performing direct translation on retrieved captions. On COCO-DEV, we see that \model only outperforms these results on English. Notice, however, that the references on COCO-DEV for the other languages rely on the M2M-100 distributions, as the baseline, promoting to an inequitable CIDEr. When evaluating on human-labeled data, as is the case with the XM3600 dataset, we see the benefits of prompting with retrieval information. 

Notice also both \model and the retrieval baseline outperform the BB model (the later also competitive to the other 3 SOTA variants), despite training with large-scale multimodal machine translated data for hours. This shows the clear benefits of using retrieval-augmentation in multilingual image captioning, not just for result quality but to avoid high computation costs as well. 

\begin{table}[!h]
    \resizebox{\linewidth}{!}{
    \centering
    \begin{tabular}{lllll}
    \toprule
        Model & en & es & hi & zh   \\ 
          \midrule    
        \multicolumn{5}{c}{COCO-DEV} \\ 
         \midrule    
         \model &  \textbf{0.767} & 0.454 & 0.334 & 0.584 \\ %
          Baseline M2M-100 &  0.590 &  0.563 & 0.548 & 0.714\\ 
            \midrule    
        \multicolumn{5}{c}{XM3600} \\ 
          \midrule    
        \model  & \textbf{0.452}  &  \textbf{0.329}  &  \textbf{0.132} &   \textbf{0.221} \\
         Baseline M2M-100 & \underline{0.333} &  \underline{0.205} & \underline{0.120} & \underline{0.170}\\
         \emph{BB}: COCO-35L & 0.297 &0.194 & 0.098 & 0.087  \\
    \bottomrule
    \end{tabular}
    }
    \caption{Comparison to direct translation on retrieved captions (Baseline), on COCO-DEV and XM3600. %
    }
    \label{tab:baseline}
\end{table}

\paragraph{Qualitative Results:}

Figure \ref{fig:examples} shows examples of captions generated in different languages by \model, together with the retrieved captions that are provided in the prompt regarding each \textit{blind-input} image. Qualitative examples tend to show diversity in the generation across the languages, with the retrieved information being itself diverse. For instance, in the first example, for English and Spanish, \model focuses on describing that a man is in front of microphones (i.e., based on the first two retrieved captions). In turn, for Hindi and Chinese, the man is in front of a laptop (i.e., from the first example), and the captions can also mention that he is ready to give a speech in Chinese (i.e., given the last two retrieved captions). In the second image, we can see that \model can simply copy a retrieved caption to generate in English, while for the other languages the model may come up with terms not directly present in the retrieved captions (e.g., ``snow slope'' in Spanish). The last image is a negative example, where incorrect retrieved captions led the model into errors in English and Chinese, showing that there are also limitations in our \textit{image-blind} approach. For more examples, see Appendix \ref{app:qual_exam}.

\section{Conclusions}

This paper proposes \model, an \textit{image-blind} few-shot multilingual image captioning model. \model is based on prompting a language model with $N$-shot examples and retrieved captions extracted by a vision-and-language model, to condition caption generation in a desired language with a multilingual language model. On XM3600, i.e. a human-labelled massively multilingual multimodal benchmark, \model performs competitively against the state-of-the-art without involving expensive training with large-scale translated multimodal data, or with any captioning data. Experimental results further demonstrate that \model largely outperforms Socratic Models \cite{zeng2022socratic}, showing that retrieval augmentation plays a crucial role in our prompting approach. As future work, we plan to further assess the use of multilingual data in the datastore, as well as the impact of directly promoting  diversity \cite{ye2022complementary,levy2022diverse} in the captions used in the prompt.

\section*{Acknowledgements}
This research was supported by the Portuguese Recovery and Resilience Plan through project C645008882-00000055, through Funda\c{c}\~ao para a Ci\^encia e Tecnologia (FCT) with the Ph.D. scholarship 2020.06106.BD, and through the INESC-ID multi-annual funding from the PIDDAC programme (UIDB/50021/2020). 

\section*{Limitations}

Image captioning and multilingual image captioning studies tend to focus on the COCO dataset, which was shown to contain gender imbalance. Previous research has also showed that models trained on COCO tend to amplify this bias \cite{hendricks2018women,zhao2017men}. While our model is not trained on COCO or in any captioning data, it relies on a pre-trained language model, which is known to suffer from different sources of bias and fairness issues \cite{bommasani2021opportunities, sheng2021societal, schramowski2022large}.  

Our model also involves retrieval-augmentation with captions extracted by a vision-and-language model, also pre-trained in an unsupervised manner. Like in the case of other retrieval-augmented generative models \cite{rag}, \model has inherently a bias towards the retrieved information. Notwithstanding, by conditioning on information from a datastore with clean and curated text, \model has potential to ameliorate some of the generation issues of the language model (e.g., elude hateful or violent language). To have insights on the biases presented in \model, we recommend analysing the retrieved captions used by the model, since they provided cues to the predictions, as shown in Figure \ref{fig:examples}. We argue that it can be much harder to have a direct interpretation for captioning models that are not retrieval-augmented. 

Another limitation of our model relates to it following a full \textit{image-blind} approach, which heavily depends on information from similar captions instead of the visual content from the actual input image. To address this limitation, future work could additionally include concepts extracted from the image in the prompt, as proposed in Socratic Models, combined with the retrieved information.

\section*{Ethics Statement}

The datasets supporting the evaluation of \model are publicly available for academic purposes. We also plan to release our code, and the additional resources that were built to support the experiments.

We emphasise that \model challenges the efficiency of most current captioning approaches, in terms of resource usage and development/deployment effort, while at the same time promoting more  equitability and inclusion, exemplified here by attempting to balance language representation at low computational costs.

We further note that while our model attempts to advance research beyond English-centric captioning, by considering captioning for a wide variety of languages, it is important to address and pay more attention to low-resource languages as well (i.e., languages beyond those covered in our tests). Evaluating \model with additional datasets, covering an even larger set of languages and concepts, would be desirable.

\bibliography{egbib,anthology,custom}
\bibliographystyle{acl_natbib}

\appendix

\section{Standard Evaluation Metrics}
\label{app:eval_metrics}

In the paper, comparison between models is performed using the CIDEr metric by following \citet{thapliyal2022crossmodal}. For holistic captioning evaluation, we provide here the performance of \model on XM3600 across additional standard automatic metrics. Specifically, Table \ref{tab:all_metrics} reports performance with BLEU-1 (B-1) and BLEU-4 (B-4) \cite{papineni2002bleu}, ROGUE-L \cite{lin2004rouge} and METEOR \cite{denkowski2014meteor}. 

\begin{table}[h!]
    \centering
    \begin{tabular}{lccccc}
    \toprule
        & B-1 &  B-4 & ROGUE-L & METEOR   \\ 
        \midrule    
        en &  0.387  & 0.067  & 0.299 & 0.129 \\
        es &  0.364 &  0.052 & 0.256 &  0.126 \\
        hi & 0.258  &  0.015 & 0.182 &  0.220\\
        zh &  0.318 &    0.053 & 0.231 & 0.105  \\ %
    \bottomrule
    \end{tabular}
    
    \caption{\model performance on the XM3600 benchmark across different evaluation metrics. 
    }
    \label{tab:all_metrics}
\end{table}

\section{Additional Results on COCO}
\label{app:more_coco}

Table \ref{tab:more_coco} provides additional results on COCO, comparing \model against against other prompt-based captioning models that do not involve training, including two previously proposed zero-shot captioning methods that are English-specific (i.e., ZeroCap \cite{tewel2021zero} and ConZic \cite{zeng2023conzic}). Results show that \model outperforms both. We also notice that unlike these models, \model works for the multilingual setting, advancing research beyond English-centric captioning.

 \begin{table}[ht!]
    \resizebox{\linewidth}{!}{
    \centering
    \begin{tabular}{lccc}
    \toprule
        Model & B-4 & METEOR & CIDEr   \\ 
        
         \midrule    
         ZeroCap  & 0.026  &  0.115  &  0.146 \\
         ConZIC  & 0.013  &  0.115  &  0.128 \\
        \model  & 0.199  &  0.220  &  0.759\\
    \bottomrule
    \end{tabular}
    }
    \caption{Results on the COCO test set. We compare our 
few-shot \model model against English-specific captioning models that do not involve supervising training as well. %
    }
    \label{tab:more_coco}
\end{table}

\section{More Qualitative Examples}
\label{app:qual_exam}

\begin{figure*}[!h]
    \centering
    \includegraphics[width=\linewidth]{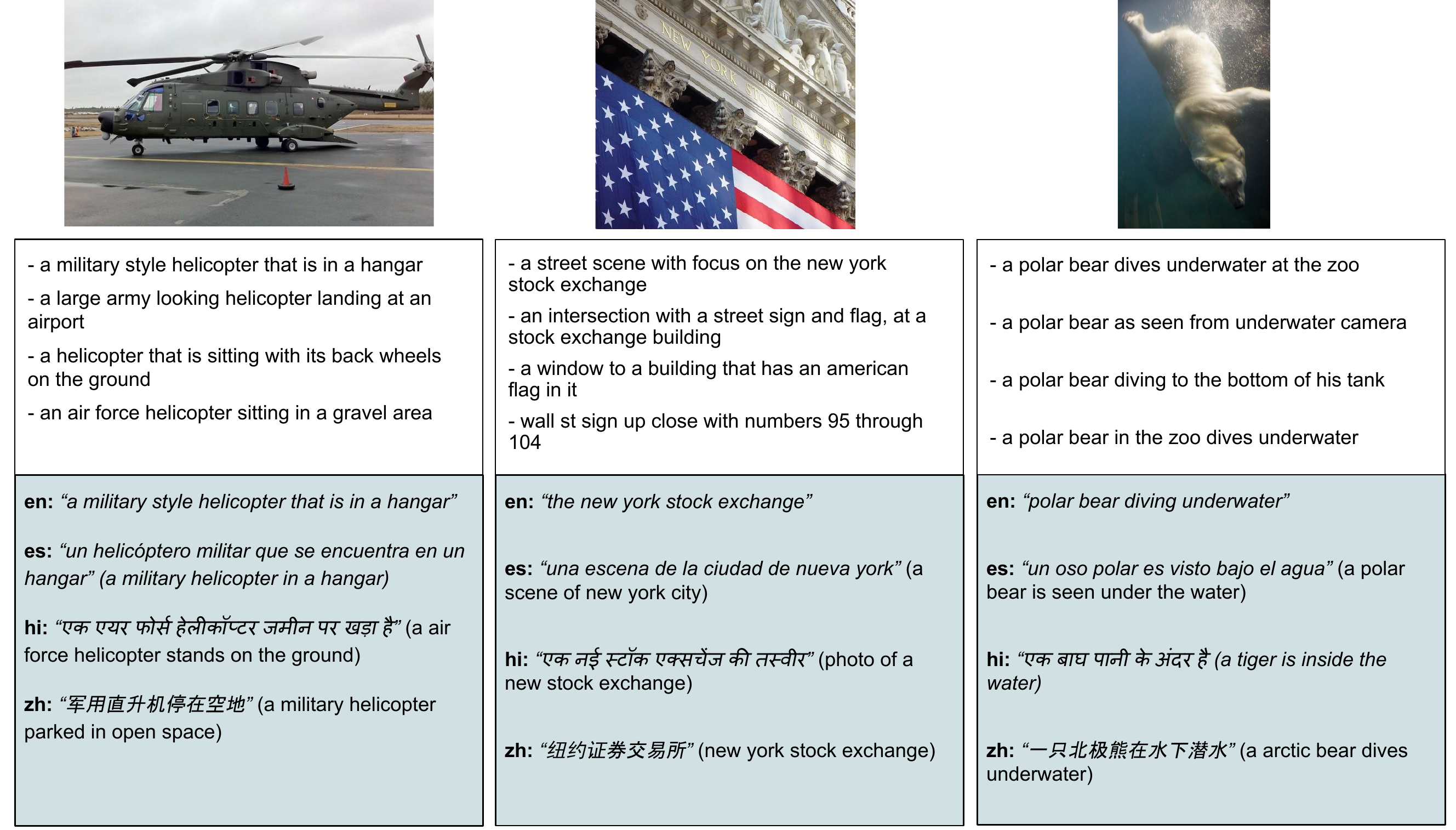}
    \caption{More examples of captions generated by \model for XM3600 images, with retrieval from COCO.}
    \label{fig:example_bear}
\end{figure*}

We provide several additional examples of captions generated from XM3600 images in Figure \ref{fig:example_bear}.

\section{Prompt-Template}
\label{app:prompt}

We follow the Socratic template, where instead of including different categories (objects, places, number of people, etc), we replace them by the retrieved captions. By following the same template, in place of a completely different one, we can assess the impact of including retrieval compared to the original Socratic framework. Our template is:
\newline

\noindent\texttt{I am an intelligent image captioning bot. Similar images have the following captions: \texttt{<caption 1>} \texttt{<caption 2>} \texttt{<caption 3>} \texttt{<caption 4>}. A creative short caption I can generate to describe this image in <language> is:}
\newline

Between the retrieved captions we use the special end of sentence token (i.e., </s>) of XGLM. Notice also that our prompt starts with 3 fixed shot examples from images in the training dataset (i.e., the same prompt is repeated multiple times to encode the n-shot examples). We share the $N$-shot examples and the set of $K$ retrieved captions used in our prompt, together with the code at \url{https://github.com/RitaRamo/lmcap}. The following text is  a concrete example of the prompt provided for the first image of XM3600.
\newline

\noindent\texttt{I am an intelligent image captioning bot. Similar images have the following captions: a horse grazing in a grassy field next to a barn</s> a brown horse grazing in its pen and a red barn and water</s> a pretty brown horse eating some grass in a bare field</s> a horse is eating grass next to a barn in the middle of a pasture</s> A creative short caption I can generate to describe this image in spanish is: Un caballo marrón es grasa cerca de una casa roja</s> I am an intelligent image captioning bot. Similar images have the following captions: a teal toilet is the center of this bathroom photo</s> a small bathroom with brightly painted blue walls</s> the bathroom has a splash of color with the blue tiles</s> the sink is above a turquoise tile sink</s> A creative short caption I can generate to describe this image in spanish is: Un baño muy limpio y bien decorado</s> I am an intelligent image captioning bot. Similar images have the following captions: a woman and child focus on a pink device in public</s> a woman holding a small child while standing near a crowd</s> a very cute lady posing with a small kid</s> a young child with a cell phone and an adult</s> A creative short caption I can generate to describe this image in spanish is: Una mujer se acercó a mirar en su teléfono mientras está listo para tomar una foto</s> I am an intelligent image captioning bot. Similar images have the following captions: a brown chicken is walking around outside with another hen</s> a couple of roosters standing in a field</s> a hen pecks the ground while another looks off in the distance</s> a couple of roosters are in a field</s> A creative short caption I can generate to describe this image in spanish is:}.

\section{Augmented Datastore Examples}
\label{app:datastore_Examples}

\begin{figure*}[!h]
    \centering
    \includegraphics[width=\linewidth]{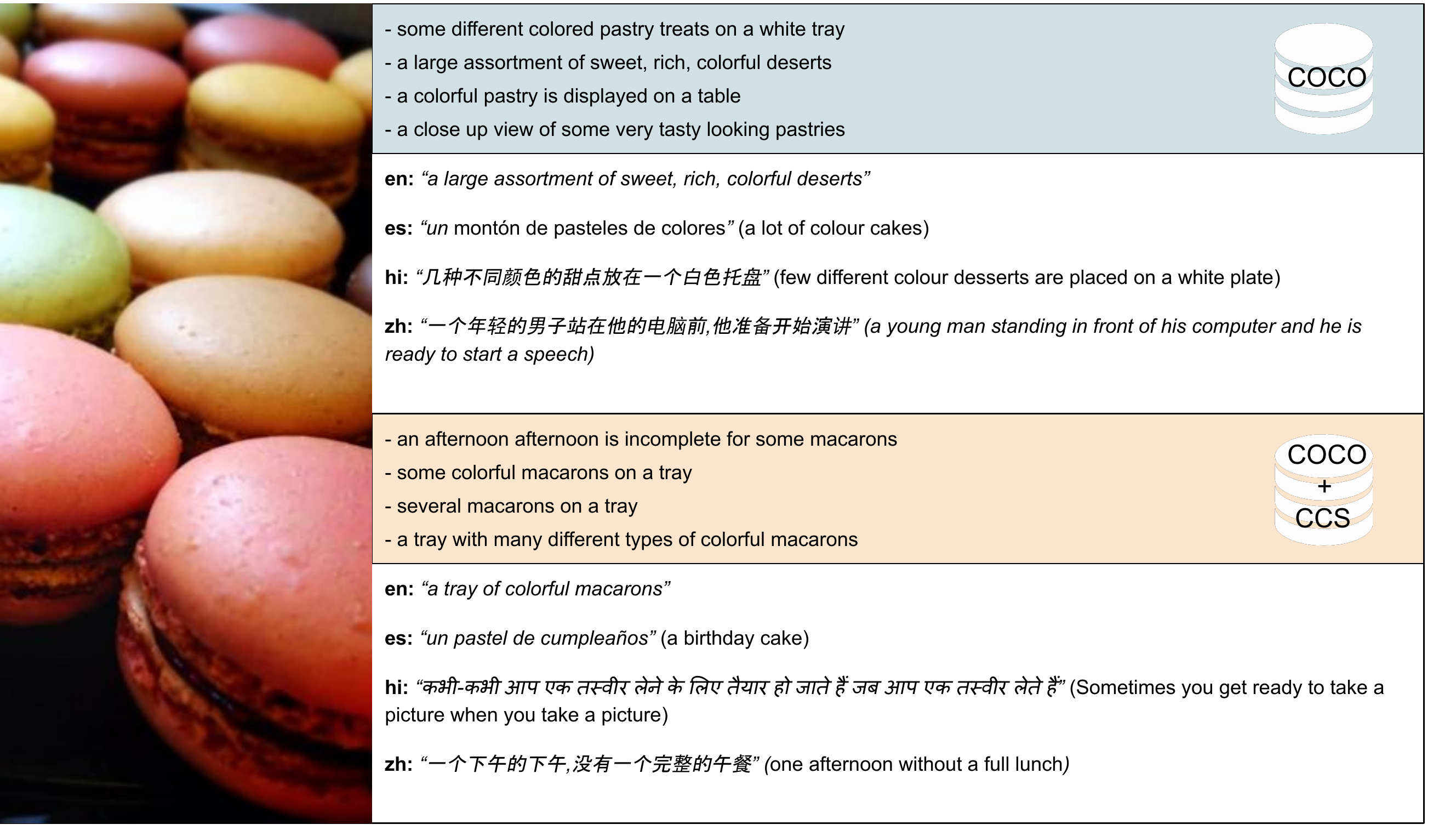}
    \caption{An example of captions generation by \model conditioned on captions retrieved from COCO (top) compared to augmenting the datastore with CCS (bottom).}
    \label{fig:datastore_examples}
\end{figure*}

\begin{figure*}
    \centering
    \includegraphics[width=\linewidth]{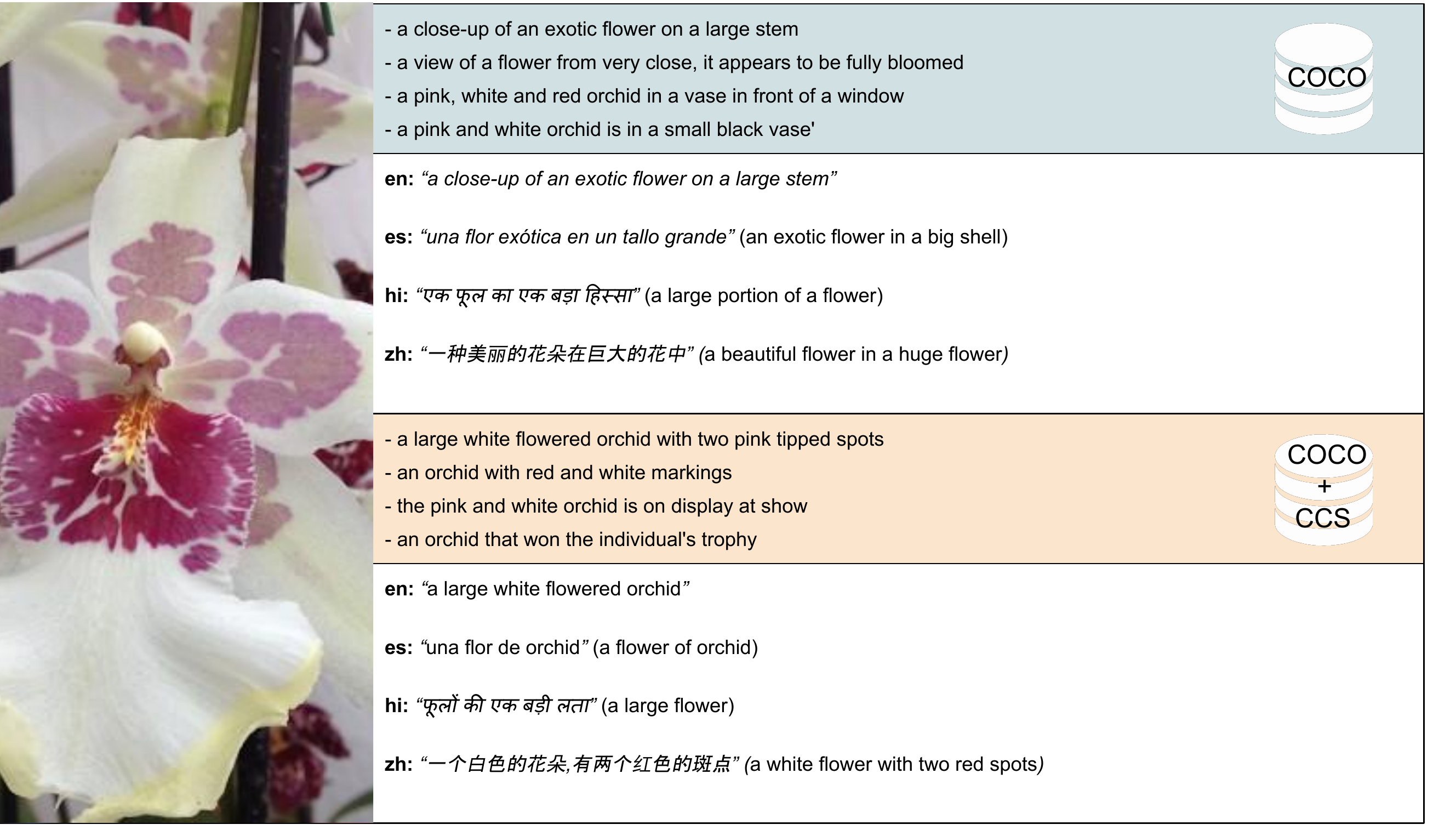}
    \caption{An example of \model generation based on retrieval from COCO (top) or COCO augmented with CCS (bottom).}
    \label{fig:datastore_examples2}
\end{figure*}

In this appendix, we provide qualitative examples on XM3600 when the datastore is augmented with CCS, i.e., with large and diverse data. In Figure \ref{fig:datastore_examples}, we can see generation improving for English, where \model correctly mentions the french concept of \textit{macarons}, available in the retrieved captions. In line with the quantitative results provided in Section \ref{ablation}, we can also see a possible explanation for why generation degraded for Hindi, that has a lower pre-training language ratio than English: \model seems to have copied the last 3-shot example provided in prompt, described above in Section \ref{app:prompt}), maybe due to presence of more noise in the CCS data. Another example can be seen in Figure \ref{fig:datastore_examples2}, where \model is more specific in generating the flower type \textit{orchid}.

\end{document}